\def\cvprPaperID{12310}
\def\confName{CVPR}
\def\confYear{2023}

\def\paperTitle{Direct Inversion: Optimization-Free Text-Driven Real Image Editing with Diffusion
Models}

\def\authorBlock{
    Adham Elarabawy\\
    UC Berkeley, Scale AI\\
    {\tt\small aelarabawy@berkeley.edu}
    \and
    Harish Kamath\\
    Scale AI\\
    {\tt\small harish.kamath@scale.com}
    \and
    Samuel Denton\\
    Scale AI\\
    {\tt\small sam.denton@scale.com}
    
}

\newif\ifreview 
\newif\ifarxiv \newcommand{\arxiv}{\arxivtrue}
\newif\ifcamera 
\newif\ifrebuttal 

\arxiv 

\pdfoutput=1
\documentclass[10pt,twocolumn,letterpaper]{article}
\ifreview \usepackage[review]{cvpr} \fi
\ifarxiv \usepackage[pagenumbers]{cvpr} \fi
\ifrebuttal \usepackage[rebuttal]{cvpr} \fi
\ifcamera \usepackage{cvpr} \fi

\usepackage{graphicx}
\usepackage{amsmath}
\usepackage{amssymb}
\usepackage{booktabs}

\usepackage{times}
\usepackage{microtype}
\usepackage{epsfig}
\usepackage[table,xcdraw]{xcolor}
\usepackage{caption}
\usepackage{float}
\usepackage{placeins}
\usepackage{color, colortbl}
\usepackage{stfloats}
\usepackage{enumitem}
\usepackage{tabularx}
\usepackage{xstring}
\usepackage{multirow}
\usepackage{xspace}
\usepackage{url}
\usepackage{subcaption}
\usepackage{xcolor}
\usepackage[hang,flushmargin]{footmisc}
\usepackage{soul}

\ifcamera \usepackage[accsupp]{axessibility} \fi

% ...

\ifarxiv  \fi

\newcommand{\R}[1]{{%
    \textbf{%
        \ifstrequal{#1}{1}{\textcolor{red}{R#1}}{%
        \ifstrequal{#1}{2}{\textcolor{blue}{R#1}}{%
        \ifstrequal{#1}{3}{\textcolor{magenta}{R#1}}{%
        \ifstrequal{#1}{4}{\textcolor{teal}{R#1}}{%
                           \textcolor{cyan}{R#1}%
        }}}}%
    }%
}}

\usepackage{xr-hyper}

\makeatletter
\newcommand*{\addFileDependency}[1]{
  \typeout{(#1)}
  \@addtofilelist{#1}
  \IfFileExists{#1}{}{\typeout{No file #1.}}
}

\makeatother

\usepackage[pagebackref,breaklinks,colorlinks]{hyperref}
\usepackage[capitalize]{cleveref}
\crefname{section}{Sec.}{Secs.}
\crefname{table}{Table}{Tables}
\crefname{figure}{Fig.}{Figs.}

\begin{document}
\title{\paperTitle}
\author{\authorBlock}

\newcommand{\eb}[1]{{\color{magenta} {\bf EB:} {\small #1}}}
\twocolumn[{
\maketitle
\begin{center}
    \captionsetup{type=figure}
    \includegraphics[width=0.95\textwidth]{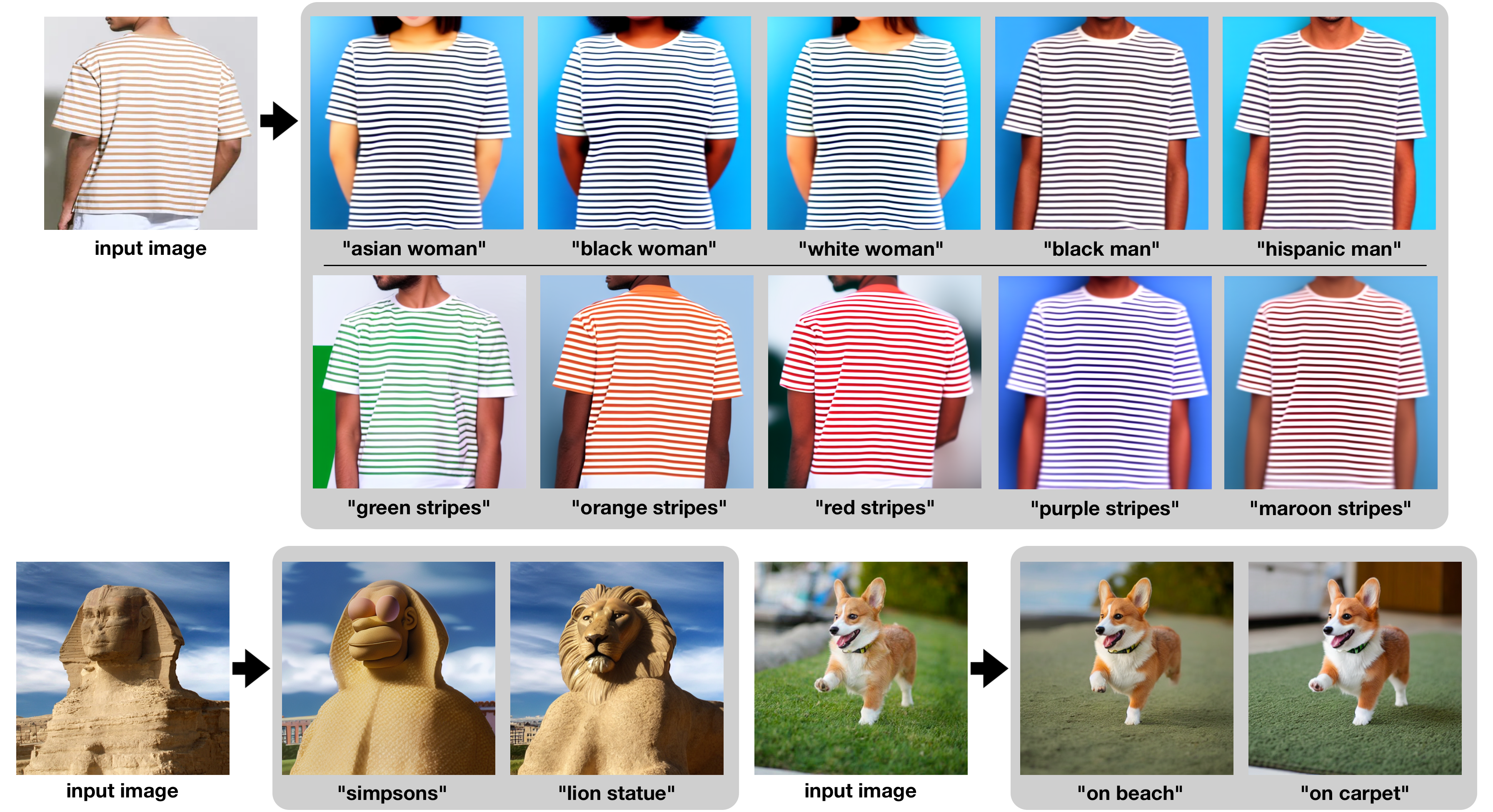}
    \captionof{figure}{\textbf{\textit{Direct Inversion} - Real image editing with no optimization or fine-tuning.} We demonstrate our method's ability to modulate pose, scene, background, style, color, and racial identity in diverse contexts in a zero-shot manner. In an effort to improve diversity of representation, we illustrate Race/Skin Tone variation with later discussion about best practices to avoid exacerbating existing biases.}
    \label{main_demo}
\end{center}
}]

\begin{abstract}
With the rise of large, publicly-available text-to-image diffusion models, text-guided real image editing has garnered much research attention recently. Existing methods tend to either rely on some form of per-instance or per-task fine-tuning and optimization, require multiple novel views, or they inherently entangle preservation of real image identity, semantic coherence, and faithfulness to text guidance. In this paper, we propose an optimization-free and zero fine-tuning framework that applies complex and non-rigid edits to a single real image via a text prompt, avoiding all the pitfalls described above. Using widely-available generic pre-trained text-to-image diffusion models, we demonstrate the ability to modulate pose, scene, background, style, color, and even racial identity in an extremely flexible manner through a single target text detailing the desired edit. Furthermore, our method, which we name \textit{Direct Inversion}, proposes multiple intuitively configurable hyperparameters to allow for a wide range of types and extents of real image edits. We prove our method's efficacy in producing high-quality, diverse, semantically coherent, and faithful real image edits through applying it on a variety of inputs for a multitude of tasks. We also formalize our method in well-established theory, detail future experiments for further improvement, and compare against state-of-the-art attempts.
\end{abstract}
\section{Introduction}
\label{sec:intro}

Manipulating real images using natural language has been a long-standing problem space in image processing. Given the wide scope of impact and potential applications, this problem space has drawn a lot of attention and research focus. As such, there have been many research attempts utilizing a wide range of methods to try and deliver robust, impressive results. Recent text-to-image machine learning models, such as Dall-E \cite{Ramesh2021ZeroShotTG}, Imagen \cite{Saharia2022PhotorealisticTD}, and Stable Diffusion \cite{Rombach2022HighResolutionIS}, have dramatically accelerated the image generation space, yielding highly coherent, diverse images that are well-aligned with text prompts. The focus of this work is to utilize these new foundational text-to-image generation models in order to \textit{edit} real images in a faithful and semantically coherent manner.

The current leading methods that attempt this research goal either (i) require significant per-instance training or fine-tuning \cite{Ruiz2022DreamBoothFT, Kim2021DiffusionCLIPTI, Gal2022AnII, Kawar2022ImagicTR, Valevski2022UniTuneTI}, (ii) are constrained to a specific domain of images \cite{Hertz2022PrompttoPromptIE, Patashnik2021StyleCLIPTM}, (iii) require secondary inputs in the form of edit-masks \cite{Avrahami2022BlendedDF, Avrahami2022BlendedLD} or multiple images of the target object \cite{Gal2022AnII, Ruiz2022DreamBoothFT}, or (iv) inherently entangle edit strength and arbitrary structural similarity\cite{Meng2021SDEditIS}, greatly limiting the types and extents of image editing possible.

We propose an image-editing technique that avoids the pitfalls above. Our method only requires an input image, along with a corresponding text prompt describing the desired edit. Given these inputs, our method is able to elaborately modify the original image using \textit{only} the target edit text prompt -- adding objects, as well as modulating scene, background, pose, style, color, and even race/ethnicity. Furthermore, \textit{Direct Inversion} is able to accomplish all of this without any sort of re-training or fine-tuning. To the best of our knowledge, ours is the first method to be able to accomplish this degree of complex image-editing without expensive per-instance or per-task fine-tuning.

At a high level, \textit{Direct Inversion} encodes the input image in the noise latent space and then "denoises" the result with CLIP text guidance, while continually injecting the initial noise back at various scales into the diffusion process. This simple formulation allows \textit{Direct Inversion} to be implemented directly into existing diffusion model implementations. In this paper, we present:

\begin{enumerate}
    \item An optimization-free, zero fine-tuning, text-based semantic image editing method that can make flexible edits to both global and local structure, attributes, style, and much more -- only requiring a single input image and text prompt.
    \item Qualitative demonstrations of the ability to perform both global style-based edits, as well as local object-level edits on real images.
    \item Quantitative demonstrations of how \textit{Direct Inversion} allows for fine-tuned control of the editability-fidelity tradeoff, as well as a technical investigation into general diffusion model inversion techniques and parameters.
    \item Ablation studies of how different parameters can affect image fidelity and edit strength using \textit{Direct Inversion}.
\end{enumerate}



\color{black}
\begin{figure}[]
    \centering
    \includegraphics[width=\linewidth]{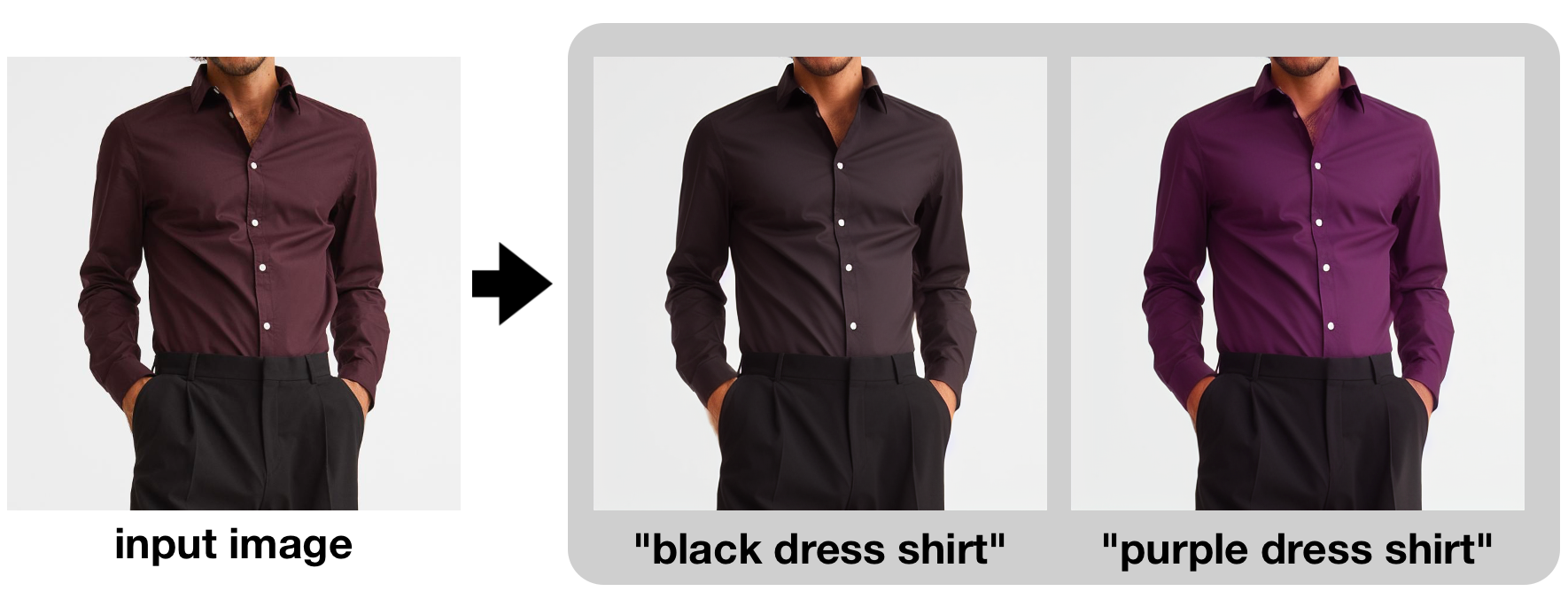}
    \caption{\textbf{Text-Guided Controlled Style Modulation.} Editing a real image using our method, \textit{Direct Inversion}, to modulate a subject's style and attributes.}
    \label{fig:shirt_demo}
\end{figure}
\section{Related Work}
\label{sec:related}

The task of image editing using generative models has been explored from multiple perspectives. In this section, we will briefly describe the most promising advances in recent years.

\textbf{GANs}

GAN\cite{GoodfellowGAN}-based approaches to the image editing task usually require a two step process. The first is to "invert" the image by being able to represent the original image in a latent space that we can then use to re-generate the original image. Once the image has been inverted, edits can be made in the latent space to re-create the adjusted image \cite{Abdal2021StyleFlowAE} \cite{Hrknen2020GANSpaceDI} \cite{Patashnik2021StyleCLIPTM} \cite{Shen2020InterpretingTL}. Recent works have improved upon this method by re-training the generative model to create images similar to the input image \cite{Alaluf2022HyperStyleSI}\cite{Bau2019SemanticPM} \cite{Roich2022PivotalTF}. However, GANs are expensive to train and suffer from generating repetitive or similar images. \cite{Karras2020TrainingGA}, making it difficult to use GANs for general image editing.

\textbf{Diffusion Models}

On the other hand, Diffusion models\cite{SohlDickstein2015DeepUL}\cite{Ho2020DenoisingDP} are more stable, but still expensive, to train and are able to produce more diversity \cite{Dhariwal2021DiffusionMB}. An advantage of diffusion models is the ability to use classifier-free guidance \cite{Ho2022ClassifierFreeDG}, which allows a user to control the output of the diffusion model without re-training the model. SDEdit \cite{Meng2021SDEditIS} requires the user to add a brush stroke to the area to edit and then de-noises the image conditioned on the desired edit — replacing the brush stroke with pixels that match the image. Other techniques such as DiffusionCLIP \cite{Kim2021DiffusionCLIPTI} utilize DDIM inversion \cite{Song2021DenoisingDI} and fine-tuning while conditioning on a CLIP-based loss during the de-noising diffusion process to bring the generated image closer to the desired edit. 

Textual Inversion \cite{Gal2022AnII} and Dream-Booth \cite{Ruiz2022DreamBoothFT} have shown strong capability in maintaining unique object characteristics while generating completely unique images. However, they require multiple images of the same object to fine-tune some parts of the diffusion model and struggle with maintaining a high level of faithfulness to the original object represented in their input images.

Finally, Imagic \cite{Kawar2022ImagicTR} presented the ability to edit a single image using a text prompt by fine-tuning a diffusion model to be faithful to the original image of interest and interpolating between the latent space representation of the original image and text prompt before performing the inversion process.

\section{Method}
\label{sec:method}

\subsection{Preliminaries}

\begin{figure*}[]
    \centering
    \includegraphics[width=\linewidth]{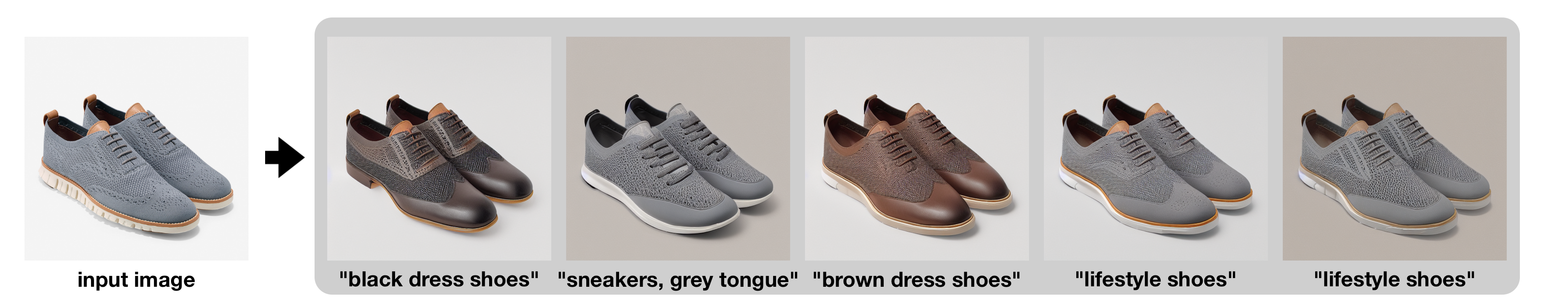}
    \caption{\textbf{Text-Guided Controlled Item Variation.} Editing a real image using our method, \textit{Direct Inversion}, to modulate an item's style/attributes.}
    \label{fig:shoes}
\end{figure*}

\subsubsection{General Diffusion Models}
Recently, diffusion models \cite{Ho2020DenoisingDP, Song2019GenerativeMB, Vincent2011ACB} have taken the generative artificial intelligence world by storm. Due to their ability to give neural networks adaptive computation time, various theoretical foundations \cite{SohlDickstein2015DeepUL, Welling2011BayesianLV, Song2019GenerativeMB, Song2021ScoreBasedGM}, relatively interpretable latent space, and unrolled process, they have become extremely powerful tools for image generation flexibly guided by various modalities of conditioning (text, image, masks, etc). They've been applied to problems of image compression, classification, restoration, and much more.

Diffusion models follow a forward and reverse process, where the forward process is a known destructive process, and the reverse process learns to undo the forward process, iteratively. In most formulations, the destructive process is a (generally stochastic) Gaussian noise perturbation. During the forward process, a clean training sample $x_0$ is iteratively corrupted over a pre-determined number of timesteps ($t \in \{0, ..., T\})$ , such that the sample at the last timestep $x_T$ is fully corrupted, containing \textit{little to no} information from the original sample $x_0$. As shown by Ho et al. \cite{Ho2020DenoisingDP}, the forward process has a neat closed-form solution:

\begin{equation}
    x_t = \sqrt{\alpha_t} x_0 + \sqrt{1 - \alpha_t} \epsilon_t
\end{equation}

When the forward process is a stochastic gaussian noise perturbation, $\epsilon_t \sim \mathcal{N}(0, \mathbf{I})$, every forward step involves sampling random noise to perturb the original sample at various noise scales. The $0 = \alpha_T < \alpha_{T-1} < ... < \alpha_1 < \alpha_0 = 1$ defines amount of noise present at each intermediate timestep, which is referred to as the noising schedule. During training, a neural network $f_\theta(x_t, t)$ takes the current sample $x_t$, and predicts the noise $\epsilon_\theta^{(t)}$. $\epsilon_\theta^{(t)}$ is then merged with $x_t$ to predict a slightly less-noisy $x_{t-1}$. The neural network objective is to make $f_\theta(x_t, t) = \epsilon_\theta^{(t)} \approx \epsilon_t$.

During inference, we start with a random sample of full noise $x_T \sim \mathcal{N}(0, \mathbf{I})$, which is refined iteratively through $t \leq T$ passes through the network. There are various sampling strategies \cite{Ho2020DenoisingDP, Song2021DenoisingDI, Liu2022PseudoNM} that define the process of merging the noise prediction $\epsilon_\theta^{(t)}$ and current (more noisy) sample $x_t$ in order to produce the previous (less noisy) sample $x_{t-1}$. The final $x_0$ sample is the resultant generated image. This procedure is therefore a learned image distribution, which has been shown to be remarkably high-fidelity and have SOTA diversity of outputs \cite{Dhariwal2021DiffusionMB}.

Additionally, diffusion models can also learn conditional distributions \cite{Ho2022ClassifierFreeDG} through the inclusion of a conditioning input $\mathbf{y}$ to the denoising network, turning $f_\theta(x_t, t)$ into $f_\theta(x_t, t, \mathbf{y})$. Given the flexibility of neural networks, the conditioning input $\mathbf{y}$ can represent various inputs: text, images, class labels, embeddings, etc. 
\subsubsection{DDIM and Determinism}
As mentioned above, there are both stochastic and deterministic ways to merge the predicted noise from the current timestep's sample $x_t$ to produce a refined estimate for the previous timestep's sample $x_{t-1}$. Song et al. proposed a method that implies determinism in the reverse process \cite{Song2021DenoisingDI}, which means that there is a direct mapping between the random noise sample $x_T$ and the clean, generated image $x_0$. Their proposed reverse process is defined as follows:

\begin{equation}
\begin{aligned}
    x_{t-1} = \sqrt{\alpha_{t-1}} \underbrace{\left(\frac{x_t - \sqrt{1-\alpha_t}\epsilon_\theta^{(t)}x_t}{\sqrt{\alpha_t}}\right)}_\text{"predicted $x_0$"} + \\\underbrace{\sqrt{1 - \alpha_{t-1} - \sigma_t^2} \epsilon_\theta^{(t)}x_t}_\text{"direction pointing to $x_t$"} + \sigma_t\epsilon_t
\end{aligned}
\end{equation}

\begin{figure*}[t]
    \centering
    \includegraphics[width=\linewidth]{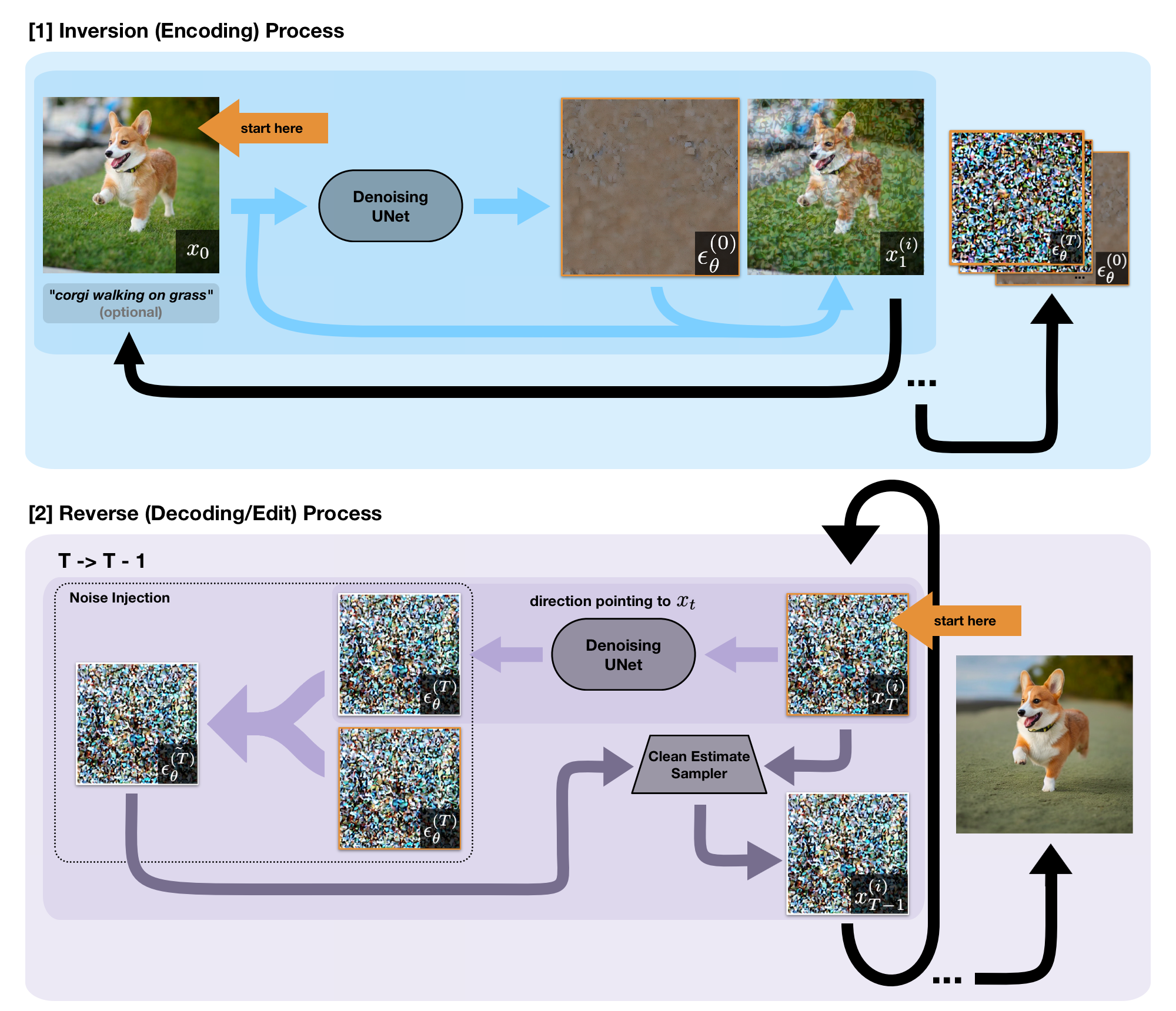}
    \caption{\textbf{Direct Inversion Process Diagram.} \textit{[1] Inversion:} First, we encode the input real image into its' encoded noises (respective to timesteps). \textit{[2] Decoding/Edit:} Then, we take the final inverted noise, pass it into the noise-prediction UNet, merge the outputted noise with the corresponding timestep's inverted noise, and then use that noise to sample the previous timestep. We repeat this process until we reach timestep 0, which corresponds to the resultant edited image.}
    \label{fig:diagram}
\end{figure*}

\begin{figure}[]
    \centering
    \includegraphics[width=\linewidth]{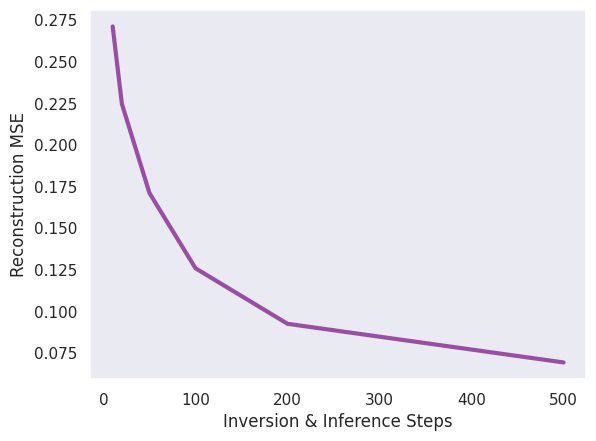}
    \caption{\textbf{DDIM Reconstruction from Inverted Latent Space.} As the number of inversion and inference steps increases, the DDIM reconstruction of the original real image becomes better.}
    \label{fig:reconmse}
\end{figure}

When $\sigma_t = 0$ for all timesteps, the reverse process becomes fully deterministic. Rather than sampling new (scaled down) noise to add to the current clean estimate, DDIM simply adds back a scaled down version of the exact noise $\epsilon_\theta^{(t)}$ predicted by the model for the current timestep. Thus, the sampled noise $x_T$ theoretically contains all of the information needed to represent the generated image $x_0$. In the context of generating new images from noise, this doesn't seem very consequential, since the same effect could be achieved by fixing the random seed for a stochastic inference process, such as in the work of Ho et al. in DDPM \cite{Ho2020DenoisingDP}. However, the closed form for this deterministic reverse process allows "working back" an existing image into its' encoded noise. Song et al. \cite{Song2021DenoisingDI} prove this theoretical framework by taking existing images $x_0$, inverting them into their noise encoding $x_T$, running the reverse process again to obtain a reconstructed $\tilde{x_0}$, then evaluating the Mean Squared Error across all reconstructed samples and the original images to quantify the information content in the noise. We reproduce their results using Stable Diffusion \cite{Rombach2022HighResolutionIS} evaluated on the CIFAR-10 test dataset \cite{Krizhevsky2009LearningML} in Figure \ref{fig:reconmse}. As expected, this figure shows that more inference and inversion steps result in decreased reconstruction loss.

\subsection{Direct Inversion}
Our method attempts to provide a no-training solution for the problem of real image editing. Given an input image $x_0$, and a text prompt $t_{edit}$ describing the desired changes, our goal is to edit the image in a coherent manner, while still remaining faithful to the identity of the original image. We accomplish this by leveraging the determinism in the forward and reverse process implied by DDIM \cite{Song2021DenoisingDI} to encode the original image $x_0$ into the noise latent space of a latent diffusion model as $\tilde{x_T}$. We then start the (deterministic) reverse process from the encoded noise, iteratively conditioning our diffusion model on the text input through classifier/classifier-free guidance and on the identity of the original image through repeated injection of the original encoded noise $\tilde{x_T}$ into the intermediate noisy timesteps for $t \in \{t-1, t-2, ..., t_{stop}\}$. The resultant method, which we name \textit{Direct Inversion}, allows for much greater control over the degree of faithfulness to the identity of the original image through modulating the magnitude and frequency of injected (encoded) noise. The use of a deterministic forward and reverse diffusion process to represent the input image in latent noise space allows our method to circumvent any optimization or fine-tuning.

As depicted in Figure \ref{fig:diagram}, our method is constituted of 2 sequential stages: (1) we invert the input image $x_0$ into its' corresponding latent noise encoding $\tilde{x_T}$; (2) we start the reverse diffusion process from $\tilde{x_T}$ for timesteps $t \in \{t, t - 1, t - 2, ..., t_{stop}\}$, conditioning on the text prompt via classifier/classifier-free guidance and the input image encoding at every step via noise injection.

\section{Experiments}
\label{sec:experiments}

\subsection{Qualitative Evaluation}
We exercise our method on a variety of real images spanning many domains in order to evaluate its' effectiveness. For each edit, we provide a concise text prompt detailing the desired edits at a very high-level. We utilize publicly available images gathered online, and curate our own target edit prompts. Given the determinism of our method, all outputs shown here are the only and primary output given a specific text prompt and real image, meaning we are not cherry picking results from a large set of outputs. This is particularly impressive for demonstrating our method's consistency when trying multiple target edits applied on a single image. Prior works leverage the one-to-many nature of their approaches in order to generate multiple outputs, which are then filtered on a per-edit approach -- sometimes resulting in up to 5X outputs produced on their end that are filtered down to the same amount we show \cite{Kawar2022ImagicTR}.

\textit{Direct Inversion} demonstrates strong results in modulating (1) pose, color, style, and structure of target objects as shown in Figure \ref{main_demo}, Figure \ref{fig:shoes}, and Figure \ref{fig:shirt_demo}, (2) background, scene, and context as shown in Figure \ref{main_demo} and Figure \ref{fig:shoes}, and (3) racial identity/skin tone/perceived gender identity of people (in an effort to improve diversity of representation) as shown in Figure \ref{main_demo}.

We note that there is implicit ambiguity in the problem formulation here, since the target edit prompt is by-definition abstract and simple. Furthermore, even if the target edit prompt is extremely specific, we might want to vary the extent to which we abide by the text prompt. Instead of deferring to diffusion models' inherent, stochastic, one-to-many nature, we instead choose to expose highly configurable hyper-parameters that modulate the strength, type, and coherence (through noise injection magnitude, frequency of noise injection, and continual injection vs single initial injection) of edit in different ways. This way, we are able to modulate target edits in a more systematic way, rather than producing many outputs and deferring to another mechanism to filter those outputs, as other methods choose to do.

\subsection{Ablation Study}
\begin{figure}[]
    \centering
    \includegraphics[width=\linewidth]{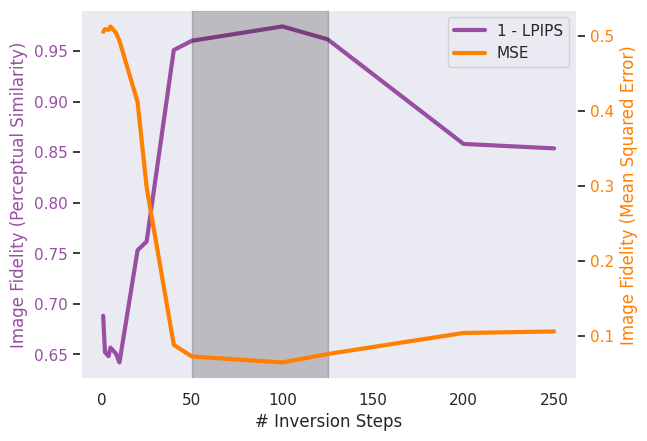}
    \caption{\textbf{Effect of Inversion Steps on Image Fidelity.} The ideal range of inversion steps is confirmed by assessing the effect on image fidelity through perceptual similarity (1 - LPIPS) and pixelwise reconstruction error (MSE). Fixed inference steps at 100. Optimal image fidelity by both similarity metrics is highlighted in grey.}
    \label{fig:inversion_ablation}
\end{figure}

\subsubsection{Effect of Inference and Inversion Steps on Image Fidelity}
Firstly, we reproduce the results shown by Song et al. \cite{Song2021DenoisingDI} in Figure \ref{fig:reconmse}, demonstrating that increasing both inversion and inference steps results in better pixel-wise reconstruction (lower MSE). Given that these results were very intuitive, we then further investigated the relationship between inversion and inference steps, and each of their effects on image fidelity, both perceptually and pixel-wise. In Figure \ref{fig:inversion_ablation}, we fix inference steps at 100 steps, and modulate the number of inversion steps $\in [0, 250]$ to see the resultant impact on perceptual similarity (1 - LPIPS \cite{zhang2018perceptual}), showing that there seems to be a region where both image fidelity by both metrics is optimal, and that naively increasing both inversion and inference steps isn't necessarily ideal. In fact, this behavior is shown to be consistent in both the perceptual and pixel-wise fidelity metrics. To the best of our knowledge, this is one of the first, if not the only, thorough investigations into the relationship between inference and inversion steps showing these conclusions.
\begin{figure}[]
    \centering
    \includegraphics[width=\linewidth]{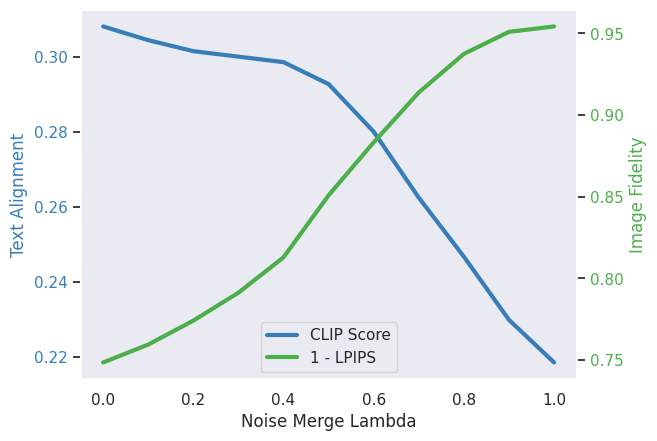}
    \caption{\textbf{Editability-Fidelity tradeoff curves.} As the scaling of the injected (inverted) noise is increased, text alignment (CLIP Similarity to prompt text) decreases, and image fidelity (LPIPS perceptual similarity to original image) increases. We performed this experiment on 215 randomized image/text pairs. Fixed Inversion/Inference steps at 100 steps each.}
    \label{fig:tradeoff}
\end{figure}

\begin{figure}[]
    \centering
    \includegraphics[width=\linewidth]{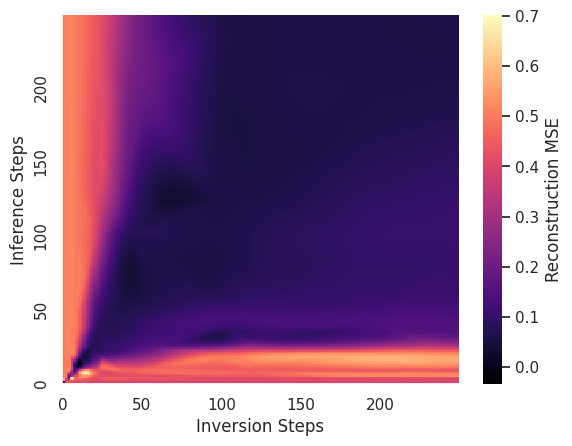}
    \caption{\textbf{Effect of Inversion Steps and Inference Steps on Image Fidelity (MSE).} A visualization of the reconstruction error landscape across parameter selections of \# inversion steps and \# inference steps. Fixed text guidance and noise injection scales.}
    \label{fig:inference_inversion_ablation}
\end{figure}

\subsubsection{Editability-Fidelity Tradeoff}
As demonstrated by Kawar and Zada et al. \cite{Kawar2022ImagicTR} and Meng et al. \cite{Meng2021SDEditIS}, it is useful to show how real-image editing methods are able to navigate the editability-fidelity tradeoff. In Figure \ref{fig:tradeoff}, we show how modulating Noise Merge Lambda (the scaling of the inverted noise that we inject into the diffusion process) affects text alignment (editability) in an inverse fashion to its' affect on image fidelity. To measure editability, we compute the CLIP similarity between the output image and the edit text prompt. To measure fidelity, we compute the perceptual similarity using LPIPS \cite{zhang2018perceptual}. Figure \ref{fig:tradeoff} clearly shows a wide range of centered feasible values, indicating a more disentangled editability-fidelity tradeoff than prior works \cite{Kawar2022ImagicTR}.

We conducted further experimentation shown in Figure \ref{fig:inference_inversion_ablation} by varying both number of inversion steps and number of inference steps and evaluating pixel-wise loss across the 2D span of parameters. The results of these experiments directed us to conclude that the parameter selection of \{\# inversion steps = 100, \# inference steps = 100\} was an optimal tradeoff between computation time and image fidelity. All figures in this paper were consequently evaluated using the above parameter selection.

\subsubsection{Effect of Text Guidance and Noise Merge Scaling on Image Fidelity}

\begin{figure*}
     \centering
     \begin{subfigure}[b]{0.45\textwidth}
         \centering
         \includegraphics[width=\textwidth]{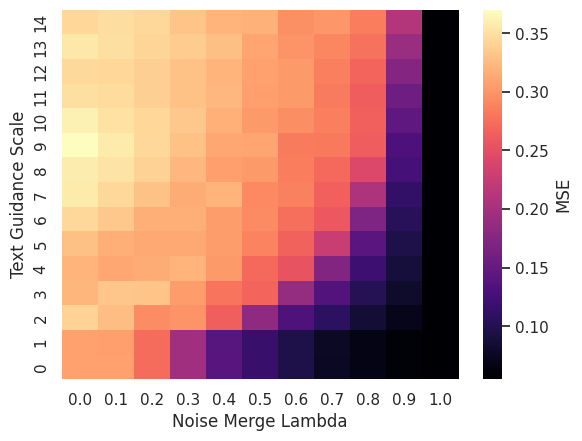}
         \caption{\textbf{Pixel-wise} Image Fidelity}
         \label{fig:mse_span}
     \end{subfigure}
     \hfill
     \begin{subfigure}[b]{0.45\textwidth}
         \centering
         \includegraphics[width=\textwidth]{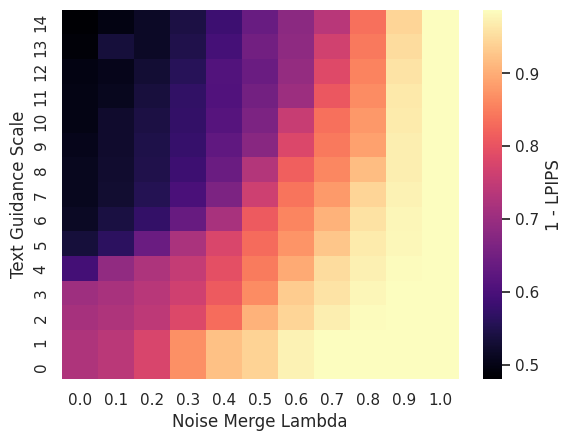}
         \caption{\textbf{Perceptual} Image Fidelity}
         \label{fig:lpips_span}
     \end{subfigure}
        \caption{\textbf{Effect of Injected Noise Scale and Text Guidance Scale on Image Fidelity.} A heatmap depicting the frontier of optimal parameter selections for text and noise guidance scales, evaluated for both perceptual and pixel losses. Fixed inversion and inference steps at 100.}
        \label{fig:spans}
\end{figure*}
\begin{figure*}[]
    \centering
    \includegraphics[width=\linewidth]{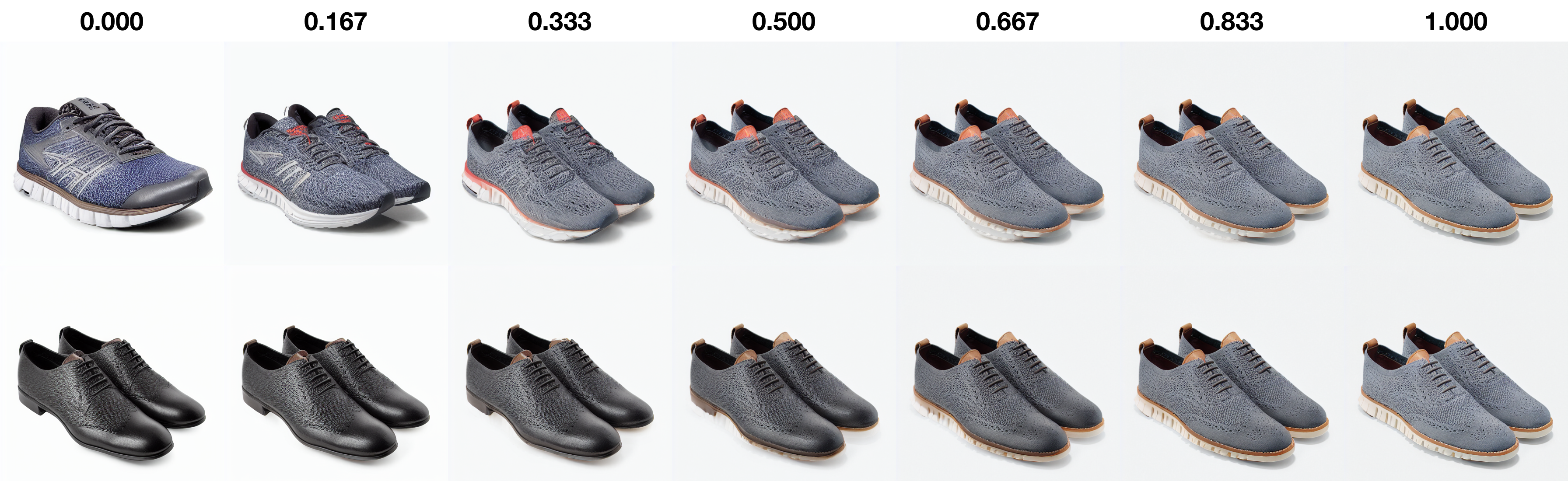}
    \caption{\textbf{Text-Guided Controlled Style Variation.} Varying noise merge scale with a fixed text guidance scale. These are visualizations of horizontal slices of Figure \ref{fig:spans}.}
    \label{fig:continuous}
\end{figure*}

Since the scale of injected noise and scale of text guidance work in opposite directions to control contrasting objectives, we conducted more image fidelity experiments, depicted in Figure \ref{fig:spans}. Along with qualitative inspection, these experiments show that there exists a frontier of feasible pairs of noise merge scale and text guidance scale. This frontier of feasible parameter selections results in image edits that balance both image fidelity and edit strength. Images that are produced through more extreme parameter selections (the extreme corners of the heatmaps) either result in the original image without any edits, or an image very faithful to the desired edit text prompt, with significant deviation from the original image. We select a few horizontal slices of Figure \ref{fig:spans} to visualise in Figure \ref{fig:continuous}. 
\section{Conclusions and Future Work}

In this work, we present \textit{Direct Inversion}, a technique for editing images that can be applied to any diffusion-based model. \textit{Direct Inversion} does not require any retraining, and can apply both global and local style changes with only the original image and a text prompt as an input. By expressing the input image in the noise latent space of the diffusion model and continually injecting it back into a reverse diffusion process, \textit{Direct Inversion} can perform image edits competitive with state-of-the-art methods in what is computationally equivalent to 2 passes through a diffusion process. Qualitative results show that \textit{Direct Inversion} is flexible in producing edits across a range of important image qualities, such as background and object-specific attributes. We additionally quantify many important measurements for image editing with \textit{Direct Inversion}, such as the editability-fidelity tradeoff, and reconstruction ability as a measure of image fidelity.

Future work should focus on further characterizing the capabilites of ``Direct Inversion" and how these capabilities are limited by the diffusion model being used. The noise latent space that \textit{Direct Inversion} acts in is rich with information that is not easily interpretable, and we believe that there is still much work to do in exploiting this space to produce new diffusion model capabilities. Furthermore, it would be useful to introduce stochasticity in part of the \textit{Direct Inversion} process to be able to generate multiple outputs for a given input.
\label{sec:conclusion}

\section{Ethics Statement}
In this work, we propose \textit{Direct Inversion}, a new technique for controlled image editing and (to a degree) synthesis. Like many other Machine Learning techniques -- especially ones focused on image synthesis --  misuse of this method could result in negative societal impacts. Although our work is created with the intent to make a positive societal impact, we acknowledge that caution and deliberation ought to be exercised in applications of these methods.

We demonstrate an ability to modulate skin tones and racial identities in existing real images. Although this ability is not unique to our presented paper (multiple recent works have similar capabilities, albeit take longer/require more inputs), we recognize that our work makes this further accessible. Additionally, it is problematic to rely on large-scale pre-trained diffusion models’ understanding of race, as this could exacerbate existing racial biases that were evident in the training of the models. Additionally, these models were trained on open source data that often include problematic and biased labels which may impact the output of our method. We implore readers and fellow researchers to continue research in a direction for systematic equality and fairness across many axes.
\label{sec:ethics_statement}

{\small
\bibliographystyle{ieee_fullname}
\bibliography{11_references}
}

\ifarxiv \clearpage \appendix
\label{sec:appendix}

 \fi

\end{document}


\title{\paperTitle \\ Supplemental Material}
\author{\authorBlock}
\maketitle

\appendix
\label{sec:appendix}


{\small
\bibliographystyle{ieee_fullname}
\bibliography{11_references}
}